\documentclass[11pt,a4paper]{article}
\usepackage{acl2023}
\usepackage{amsmath}
\usepackage{amssymb}
\usepackage{graphicx}
\usepackage{booktabs}
\usepackage{multirow}
\usepackage{tabularx}
\usepackage{array}
\usepackage{caption}
\usepackage{subcaption}
\usepackage{url}
\usepackage{xcolor}

\title{Caption Generation for Dongba Paintings via Prompt Learning and Semantic Fusion}

\author{Shuangwu Qian \quad Xiaochan Yuan \quad Pengfei Liu$^*$ \\
  Sichuan Agricultural University \\ 
 \texttt{\{shuangwuqian, xiaochanyuan, pengfeiliu\}@stu.sicau.edu.cn}}

\date{}

\begin{document}
\maketitle

\begin{abstract}
Dongba paintings, the treasured pictorial legacy of the Naxi people in southwestern China, feature richly layered visual elements, vivid color palettes, and pronounced ethnic and regional cultural symbolism, yet their automatic textual description remains largely unexplored owing to severe domain shift when mainstream captioning models are applied directly.
This paper proposes \textbf{PVGF-DPC} (\textit{Prompt and Visual Semantic-Generation Fusion-based Dongba Painting Captioning}), an encoder--decoder framework that integrates a content prompt module with a novel visual semantic-generation fusion loss to bridge the gap between generic natural-image captioning and the culturally specific imagery found in Dongba art.
A MobileNetV2 encoder extracts discriminative visual features, which are injected into the layer normalization of a 10-layer Transformer decoder initialized with pretrained BERT weights; meanwhile, the content prompt module maps the image feature vector to culture-aware labels---such as \emph{deity}, \emph{ritual pattern}, or \emph{hell ghost}---and constructs a post-prompt that steers the decoder toward thematically accurate descriptions.
The visual semantic-generation fusion loss jointly optimizes the cross-entropy objectives of both the prompt predictor and the caption generator, encouraging the model to extract key cultural and visual cues and to produce captions that are semantically aligned with the input image.
We construct a dedicated Dongba painting captioning dataset comprising 9{}408 augmented images with culturally grounded annotations spanning seven thematic categories. Extensive experiments demonstrate that PVGF-DPC attains BLEU-1/2/3/4 scores of 0.603/0.426/0.317/0.246, a METEOR score of 0.256, a ROUGE score of 0.403, and a CIDEr score of 0.599 on the Dongba painting test set, substantially outperforming state-of-the-art zero-shot and controllable captioning baselines including BLIP, ViECap, MacCap, and ClipCap in both objective metrics and subjective cultural expressiveness.
\end{abstract}

\section{Introduction}

Image captioning---the task of automatically generating a natural-language sentence that faithfully describes the visual content of a given image---has attracted sustained interest as a canonical problem at the intersection of computer vision and natural language processing~\cite{vinyals2015show,hossain2019comprehensive,stefanini2022show,wang2022image}. Beyond enumerating the objects present in a scene, a high-quality caption must articulate the relationships among entities, their attributes, and their activities in a coherent linguistic form~\cite{karpathy2015deep}. To date, image captioning has found applications in assistive technology for the visually impaired, intelligent human--computer interaction, robotics~\cite{tang2017image,waghmare2022image}, visual question answering, and medical image annotation, among others. Recent advances in cross-modal video reconstruction~\cite{huang2025mindev}, and multimodal video understanding benchmarks~\cite{nie2025chinesevideobench} have further demonstrated that modern neural architectures can reliably encode and decode structured information across heterogeneous modalities, underscoring the potential for extending captioning paradigms to underexplored visual domains.

Dongba paintings constitute a treasured form of Naxi pictorial art originating in the Yunnan--Tibet border region of southwestern China. Executed primarily in bold line work with vivid and contrasting colors, these paintings serve as a visual compendium of Dongba religion, mythology, and daily life~\cite{qian2020simulation}. As illustrated in Figure~1, each painting embodies rich cultural connotations: tigers symbolize valor and serve as guardian beasts of the Dongba altar; deities are depicted seated upon lotus thrones surrounded by radiant halos; and the recurring \emph{Eight Treasures of the Naxi}---the purification vase, the endless knot, the conch shell, the lotus, the twin fish, and others---encode aspirations for auspicious fortune and spiritual purity. Despite the profound cultural significance of these works, research on Dongba paintings has hitherto concentrated on the analysis of artistic value~\cite{yang2021study} and stylistic evolution~\cite{su2022blending}, with virtually no prior effort devoted to generating textual descriptions that capture both the visual content and the underlying cultural semantics.

The evolution of image captioning has proceeded through several paradigmatic stages. Early approaches relied on template-based methods~\cite{li2011composing} or nearest-neighbor retrieval~\cite{devlin2015exploring}, which, while straightforward, produce repetitive descriptions tied to rigid syntactic structures. The deep-learning era was inaugurated by the Show-and-Tell model~\cite{vinyals2015show}, which employs a convolutional neural network (CNN) as a visual encoder and a long short-term memory (LSTM)~\cite{hochreiter1997long} network as a language decoder. Subsequent work introduced attention mechanisms~\cite{xu2015show,anderson2018bottomup}, wherein the decoder dynamically attends to salient image regions, wherein the decoder dynamically attends to salient image regions, enabling the decoder to selectively attend to salient image regions and thereby generate more detailed captions. More recent investigations have explored self-critical sequence training~\cite{rennie2017self,ren2017deep}, memory-augmented transformers~\cite{cornia2020meshed}, and deep visual-semantic alignment~\cite{karpathy2015deep}, progressively improving both fluency and fidelity. Meanwhile, vision--language pre-training paradigms such as CLIP~\cite{radford2021learning}, BLIP~\cite{li2022blip}, BLIP-2~\cite{li2023blip2}, OFA~\cite{wang2022ofa}, GIT~\cite{wang2022git}, Oscar~\cite{li2020oscar}, and VinVL~\cite{zhang2021vinvl} have demonstrated remarkable zero-shot and few-shot captioning capabilities by jointly learning visual and textual representations at scale~\cite{jia2021scaling}. These models, however, tend to produce generic descriptions and struggle to capture the distinctive stylistic and thematic features characteristic of specialized visual domains.

To address the need for more targeted generation, the paradigm of \emph{controllable image captioning} (CIC) has emerged, wherein auxiliary control signals are introduced to steer the output toward desired attributes. Zheng et al.~\cite{zheng2019intention} propose the CGO model, which selects a guiding word as an anchor and predicts the surrounding context bidirectionally via an LSTM. Wang et al.~\cite{wang2023controllable} embed prompt learning into the captioning framework, using prompt tokens to elicit stylized captions. Fei et al.~\cite{fei2023transferable} introduce ViECap, which obtains entity-aware hard prompts through grammar parsing and combines them with a masking strategy to direct the language model's attention toward entity-level information. Hu et al.~\cite{hu2023promptcap} introduce PromptCap, which takes a natural language prompt to control the visual entities described in the generated caption, enabling task-aware captioning for downstream VQA. Hu et al.~\cite{hu2023promptcap} introduce PromptCap, which takes a natural language prompt to control the visual entities described in the generated caption, enabling task-aware captioning for downstream VQA. Qiu et al.~\cite{qiu2024mining} present MacCap, which retrieves keywords from a memory bank and uses them to condition caption generation. Additionally, recent investigations into intent understanding under ambiguous conditions~\cite{he2025enhancing}, coherent narrative generation with retrieval enhancement~\cite{yi2025score}, agentic context engineering for long-document comprehension~\cite{liu2025resolving}, have highlighted the importance of guiding generative models with structured, context-aware prompts---a principle that lies at the heart of our approach.

Despite the progress outlined above, directly applying existing captioning methods to Dongba paintings introduces two critical challenges. First, deep-learning models typically require large-scale, high-quality image--caption pairs for effective training; the scarcity of Dongba painting samples and the insufficiency of associated descriptive annotations lead to overfitting~\cite{shorten2019survey}~\cite{shorten2019survey}, degrading generalization performance. Second, a pronounced \emph{domain shift} problem arises: pre-trained models carry strong priors learned from natural-image distributions and, when confronted with the stylized and culturally charged imagery of Dongba art, they tend to generate hallucinated or culturally irrelevant descriptions. Existing controllable captioning models largely rely on extracting explicit keywords or entities from the text; they cannot capture the implicit cultural connotations and layered semantic depth inherent to Dongba paintings, resulting in superficial and one-dimensional descriptions.

To overcome these limitations, this paper proposes PVGF-DPC, a captioning framework specifically designed for the Dongba painting domain. The principal contributions of this work are as follows:

\begin{enumerate}
 \item \textbf{A dedicated Dongba painting captioning dataset.} We collect authentic Naxi Dongba paintings from prior literature~\cite{qian2020simulation} and organize the images into seven thematic categories---deity and spirit, hell ghost, bird and beast, flora, horseback riding and fishing, music and dance, and religious pattern---accompanied by culturally grounded textual descriptions. Data augmentation expands the dataset to 9{}408 images.
 \item \textbf{A content prompt module.} Leveraging the image feature vector, this module identifies the subject, action, and thematic category of the painting, and constructs a post-prompt that furnishes the decoder with explicit cultural context, thereby mitigating hallucination and enhancing thematic relevance.
 \item \textbf{A visual semantic-generation fusion loss.} By jointly optimizing the cross-entropy losses of both the prompt predictor and the caption generator, this loss encourages the encoder to extract culturally salient visual features and the decoder to produce content-centric, semantically faithful descriptions.
\end{enumerate}

Extensive experiments on the Dongba painting test set show that PVGF-DPC achieves state-of-the-art results across all seven evaluation metrics, surpassing competitive baselines by substantial margins and delivering qualitatively superior descriptions that preserve the ethnic and cultural nuances of the source imagery.

\section{Related Work}

\subsection{Image Captioning}

Image captioning has undergone a paradigm shift from early template-based and retrieval-based approaches~\cite{li2011composing,devlin2015exploring} to end-to-end neural architectures that jointly encode visual and linguistic information. The foundational encoder--decoder framework proposed by Vinyals et al.~\cite{vinyals2015show} pairs a CNN image encoder with an LSTM language decoder, and remains the conceptual backbone of most subsequent models. The introduction of the visual attention mechanism by Xu et al.~\cite{xu2015show} enables the decoder to dynamically focus on relevant image regions at each generation step, significantly improving descriptive detail. Anderson et al.~\cite{anderson2018bottomup} further advance this line of work with a bottom-up and top-down attention architecture that extracts object-level features via Faster R-CNN and re-weights them through a top-down attention LSTM, achieving state-of-the-art results on the MSCOCO benchmark~\cite{chen2015microsoft}.

Parallel research has explored diverse strategies for improving caption quality. Rennie et al.~\cite{rennie2017self} propose self-critical sequence training (SCST), which employs reinforcement learning with the model's own greedy output as the baseline reward, substantially boosting CIDEr scores. Cornia et al.~\cite{cornia2020meshed} introduce the Meshed-Memory Transformer, which replaces recurrent decoders with a fully attentive architecture and augments it with learned memory vectors, enabling richer cross-layer interactions. Yao et al.~\cite{yao2018exploring}, and Chen et al.~\cite{chen2024expanding} incorporate visual relationship modeling by detecting inter-object relations, thereby generating captions that articulate spatial and semantic dependencies. Johnson et al.~\cite{johnson2016densecap} introduce DenseCap, which jointly trains object detection and description generation for dense region-level captioning. Luo et al.~\cite{luo2018discriminability} design a discriminability objective that encourages captions to distinguish the target image from distractor images, promoting specificity over generic descriptions. Meanwhile, the proliferation of vision--language pre-training models---including CLIP~\cite{radford2021learning}, BLIP~\cite{li2022blip}, BLIP-2~\cite{li2023blip2}, OFA~\cite{wang2022ofa}, and GIT~\cite{wang2022git}---has shifted the landscape toward zero-shot and few-shot captioning, wherein models pre-trained on web-scale image--text pairs are adapted to novel domains with minimal or no task-specific supervision~\cite{cho2021unifying,zhou2020unified}.

Within the zero-shot captioning paradigm, several noteworthy approaches have emerged. Li et al.~\cite{li2023decap} introduce DeCap, which decodes CLIP latent representations into captions via text-only training, circumventing the need for paired image--caption data. Mokady et al.~\cite{mokady2021clipcap} propose ClipCap, which learns a lightweight mapping from CLIP visual features to a prefix that conditions GPT-2 generation. Yu et al.~\cite{yu2023cgtgan} present CgT-GAN, a CLIP-guided text GAN that generates diverse captions through adversarial training. Wang et al.~\cite{wang2022git} design GIT, a generative image-to-text transformer that unifies captioning and VQA under a single architecture with a simplified image encoder--text decoder design. Wang et al.~\cite{wang2022git} design GIT, a generative image-to-text transformer that unifies captioning and VQA under a single architecture with a simplified image encoder--text decoder design. Zeng et al.~\cite{zeng2024meacap} introduce MeaCap, a memory-augmented approach that retrieves and integrates relevant textual fragments from an external memory bank during generation. Despite their flexibility, competitive benchmarks such as the NICE zero-shot captioning challenge~\cite{kim2023nice} have further revealed that competitive benchmarks such as the NICE zero-shot captioning challenge~\cite{kim2023nice} have further revealed that these zero-shot models uniformly produce generic descriptions and lack the capacity to convey domain-specific cultural nuances, a limitation that motivates the culturally grounded approach proposed in this paper. Recent surveys on deep-learning-based captioning~\cite{zhao2023deep} further emphasize the need for domain-adaptive strategies when applying general-purpose models to specialized visual domains.

Recent advances in multimodal document understanding---including universal large multimodal models for document analysis~\cite{feng2023unidoc,feng2024docpedia}, text-centric visual question answering~\cite{tang2024mtvqa,tang2024textsquare}, and vision--language surveys~\cite{wang2025vision,wang2025wilddoc}---have demonstrated the power of aligning visual and textual modalities at fine granularity. Scene text detection and recognition methods~\cite{tang2022few,tang2022optimal,tang2022youcan,tang2023character,liu2023spts,li2024real,lu2024bounding} further underscore the importance of precisely localizing and interpreting textual cues within complex visual scenes, a core capability that parallels the need for accurate subject identification in Dongba painting captioning.

\subsection{Controllable Image Captioning}

Controllable image captioning extends the standard captioning paradigm by introducing additional control signals---such as style, sentiment, length, or content constraints---to guide the generation process~\cite{zheng2019intention,wang2023controllable,stefanini2022show}. Early work in this direction focuses on style transfer, where captions are generated in a specified linguistic style (e.g., humorous, romantic, or formal) while preserving factual accuracy~\cite{luo2018discriminability}. More recently, prompt-based approaches have gained prominence. The success of large language models~\cite{brown2020language} has catalyzed the development of large multimodal models such as Flamingo~\cite{alayrac2022flamingo}, LLaVA~\cite{liu2023llava}, and InstructBLIP~\cite{dai2023instructblip} have demonstrated that instruction-tuned vision--language architectures can follow diverse user prompts to generate contextually appropriate descriptions. The success of large language models~\cite{brown2020language} has catalyzed the development of large multimodal models such as Flamingo~\cite{alayrac2022flamingo}, LLaVA~\cite{liu2023llava}, and InstructBLIP~\cite{dai2023instructblip} have demonstrated that instruction-tuned vision--language architectures can follow diverse user prompts to generate contextually appropriate descriptions. The power of scale for parameter-efficient prompt tuning demonstrated by Lester et al.~\cite{lester2021power} has inspired a wave of methods that inject learnable prompt tokens into pre-trained language models to steer generation without full fine-tuning~\cite{liu2021pre}. In the captioning domain, Wang et al.~\cite{wang2023controllable} integrate prompt learning into the captioning pipeline, enabling the generation of stylized captions conditioned on user-defined prompts. Li et al.~\cite{li2024learning} further propose learning combinatorial prompts for universal controllable captioning, allowing multiple control signals to be composed flexibly. Li et al.~\cite{li2024learning} further propose learning combinatorial prompts for universal controllable captioning, allowing multiple control signals to be composed flexibly. Fei et al.~\cite{fei2023transferable} propose ViECap, which employs grammar parsing to extract entity-aware hard prompts and uses a masking strategy to focus the language model on entity-level information, thereby improving the factual grounding of generated captions. Qiu et al.~\cite{qiu2024mining} develop MacCap, which leverages a filtering retrieval module to obtain image-relevant keywords from a memory bank and uses these keywords to condition the captioning process.

Large-scale foundation models such as Dolphin-v2~\cite{feng2026dolphinv2} have demonstrated that hierarchical visual parsing and multi-granularity feature alignment can substantially improve content grounding in generative models. Complementary work on lightweight temporal-aware video editing, autonomous decision-making with vision--language guidance~\cite{wang2025pargo}, and benchmark construction for multimodal cognition~\cite{shan2024mctbench} further highlights the versatility of prompt-driven architectures across diverse vision--language tasks. These developments collectively motivate our design of a content prompt module that leverages culture-aware labels as structured prompts, steering the decoder to generate Dongba painting descriptions that are both visually grounded and culturally informed.

\subsection{Cultural Heritage Image Understanding}

The application of image captioning techniques to cultural heritage imagery represents a nascent but important research direction~\cite{wang2023cultural}. Unlike natural photographs, artworks and cultural artifacts exhibit distinctive stylistic conventions, symbolic vocabularies, and culturally encoded meanings that generic captioning models are ill-equipped to handle. The ArtEmis dataset~\cite{achlioptas2021artemis}, which provides emotion attributions and grounded explanations for 80K artworks, represents a pioneering effort to bridge affective understanding and visual art, yet it does not address the challenge of culture-specific iconographic description. The ArtEmis dataset~\cite{achlioptas2021artemis}, which provides emotion attributions and grounded explanations for 80K artworks, represents a pioneering effort to bridge affective understanding and visual art, yet it does not address the challenge of culture-specific iconographic description. Image style transfer methods~\cite{gatys2016image} can reproduce the visual appearance of artistic styles but do not address the semantic interpretation of cultural content. Recent benchmarking efforts for Chinese art understanding~\cite{yu2025benchmarking}, CNN-based classification of Chinese painting styles~\cite{chen2022classification},, CNN-based classification of Chinese painting styles~\cite{chen2022classification}, and comprehensive visual table understanding~\cite{zhao2025tabpedia} have revealed persistent gaps between model outputs and human-level cultural comprehension.

Dongba painting, as a unique form of Naxi pictorial expression, presents particularly acute challenges. Its iconography draws from a closed symbolic system---deities, ritual implements, mythological creatures---whose meanings are opaque to models trained on natural-image distributions. Prior computational research on Dongba art has focused on stylistic simulation~\cite{qian2020simulation} and aesthetic analysis~\cite{su2022blending,yang2021study}, but the automatic generation of culturally faithful textual descriptions has not been addressed. Harmonizing visual text comprehension and generation~\cite{zhao2024harmonizing}, multi-modal in-context learning~\cite{zhao2024multi}, and cross-modal enhancement techniques~\cite{niu2025cme,jia2025meml} suggest promising avenues for bridging the cultural gap. Grounded text-to-image generation approaches~\cite{li2023gligen} further demonstrate the value of explicit spatial and semantic grounding in visual generation pipelines. This paper bridges the gap by proposing a prompt-guided captioning model that explicitly encodes Dongba cultural knowledge into the generation process.

\section{Method}

\subsection{Overall Architecture}

The proposed PVGF-DPC framework adopts an encoder--decoder architecture augmented with a content prompt module, as illustrated in Figure~2. The pipeline comprises three principal components: (1) an image encoder that extracts a discriminative visual representation of the input Dongba painting, (2) a content prompt module that infers the thematic category and cultural attributes of the image, and (3) a Transformer-based text decoder that generates a culturally grounded caption conditioned on both the visual features and the prompt information.

Formally, given an input Dongba painting image, the image is first resized and normalized to a resolution of $299 \times 299 \times 3$ pixels. The encoder then processes the image and produces a 1{}280-dimensional feature vector $\mathbf{e}_x$. Simultaneously, the associated caption text is tokenized into a sequence of length 128. The feature vector $\mathbf{e}_x$ is injected into each coding layer of the decoder to modulate the text generation process. In parallel, $\mathbf{e}_x$ is fed into the content prompt module, which produces a prompt embedding $\mathbf{e}_p$ encoding the image's subject and action information. The prompt embedding is concatenated with the caption text embedding and passed to the decoder. The model is trained end-to-end by jointly optimizing the content prompt module and the text generation objective through the visual semantic-generation fusion loss.

\subsection{Encoder}

The encoder employs a convolutional neural network to capture the salient semantic features of Dongba paintings. We adopt MobileNetV2~\cite{sandler2018mobilenetv2} as the backbone, comparing it against VGG16~\cite{simonyan2014very}, DenseNet121~\cite{huang2017densely}, EfficientNetB0~\cite{tan2019efficientnet}, and ResNet-50~\cite{he2016deep}. While vision transformers such as ViT~\cite{dosovitskiy2021an} and Swin Transformer~\cite{liu2021swin} have demonstrated strong performance on large-scale benchmarks, their heavy computational requirements make CNN-based encoders more suitable for our limited-data Dongba painting setting (see Section~4.5), comparing it against VGG16~\cite{simonyan2014very}, DenseNet121~\cite{huang2017densely}, EfficientNetB0~\cite{tan2019efficientnet}, and ResNet-50~\cite{he2016deep}. While vision transformers such as ViT~\cite{dosovitskiy2021an} and Swin Transformer~\cite{liu2021swin} have demonstrated strong performance on large-scale benchmarks, their heavy computational requirements make CNN-based encoders more suitable for our limited-data Dongba painting setting (see Section~4.5), motivated by its favorable trade-off between feature extraction quality and computational efficiency. The inverted residual block architecture and depthwise separable convolution technique substantially reduce the model's parameter count and computational cost while maintaining expressive power.

To preserve the artistic detail and compositional characteristics of Dongba paintings, the mean height and width across the entire dataset are computed, and all images are uniformly resized to $299 \times 299$ pixels with three RGB channels. Upon entering the encoder, the image first passes through an initial convolutional layer and a standard $3 \times 3$ convolution, producing 32 output channels. This stage reduces the spatial dimensions while extracting basic visual features such as color distributions and coarse contours. The features then traverse 16 inverted residual blocks, each comprising three stages: an \emph{expansion layer} that increases the channel dimensionality to capture complex cultural elements (e.g., costume details, decorative patterns, symbolic motifs), a \emph{depthwise convolution layer} that extracts spatial features capable of recognizing character poses, facial expressions, and nuanced background gradients, and a \emph{projection layer} that compresses the feature maps to retain essential artistic elements while discarding redundant information. After all inverted residual blocks, a $1 \times 1$ convolution increases the output channel count to 1{}280, and global average pooling converts the feature map into a 1{}280-dimensional feature vector $\mathbf{e}_x$.

\subsection{Decoder}

To fully leverage the contextual information in Dongba painting captions, the decoder integrates both the image features and the prompt information to produce culturally rich textual descriptions. We employ a Transformer-based architecture~\cite{vaswani2017attention} that exploits self-attention to capture long-range dependencies within the input sequence, facilitating the generation of accurate and coherent text. The decoder is composed of 10 coding layers and is initialized with pretrained BERT~\cite{devlin2019bert} weights to enhance semantic understanding and accelerate convergence.

\paragraph{Input processing.} The output of the content prompt module is appended to the caption text as a post-prompt. For instance, for a painting depicting a deity seated upon a lotus throne, the prompt module produces the prompt text ``This is a Dongba painting about a deity,'' which is concatenated with the caption and tokenized into a token sequence $Y$. The token sequence is then projected through an embedding layer before being passed to the coding layers.

\paragraph{Coding layer structure.} Each coding layer consists of two sub-layers: a multi-head self-attention mechanism and a feed-forward network, connected via residual connections and layer normalization. The multi-head self-attention computes:
\begin{equation}
\mathrm{A}(\mathbf{Q}, \mathbf{K}, \mathbf{V}) = \mathrm{softmax}\!\left(\frac{\mathbf{Q}\mathbf{K}^\top}{\sqrt{d_k}}\right) \mathbf{V}
\label{eq:attention}
\end{equation}
where $\mathbf{Q}$, $\mathbf{K}$, and $\mathbf{V}$ denote the query, key, and value matrices, respectively, and $d_k$ is the query dimensionality.

Crucially, the image feature vector $\mathbf{I} = \mathbf{e}_x$ is integrated into the layer normalization~\cite{ba2016layer}~\cite{ba2016layer} to fuse visual and textual information. The first layer normalization with residual connection computes:
\begin{equation}
\mathbf{x}' = \gamma \cdot \frac{\mathbf{x} + \mathrm{A}(\mathbf{Q}, \mathbf{K}, \mathbf{V}) + \mathbf{I} - \mu}{\sqrt{\sigma^2 + \epsilon}} + \beta
\label{eq:layernorm1}
\end{equation}
where $\mu$ and $\sigma^2$ denote the mean and variance of $\mathbf{x} + \mathrm{A}(\mathbf{Q}, \mathbf{K}, \mathbf{V}) + \mathbf{I}$, and $\gamma$, $\beta$ are learnable scale and bias parameters, and $\epsilon$ is a small constant for numerical stability.

The output $\mathbf{x}'$ is then processed by the feed-forward network $\mathrm{FFN}(\cdot)$, which applies two linear transformations with a ReLU activation:
\begin{equation}
\mathbf{x}'' = \gamma \cdot \frac{\mathbf{x}' + \mathrm{FFN}(\mathbf{x}') + \mathbf{I} - \mu'}{\sqrt{\sigma'^2 + \epsilon}} + \beta
\label{eq:layernorm2}
\end{equation}
This feature modulation strategy, reminiscent of FiLM conditioning~\cite{perez2018film}, ensures that the visual semantics of the Dongba painting continuously guide the text generation at every layer.

\paragraph{Output and loss.} After 10 coding layers, the resulting hidden state $\mathbf{H}$ is projected to vocabulary logits via a linear layer and converted to a probability distribution through softmax. The text generation loss is the standard cross-entropy:
\begin{equation}
\mathcal{L}_{\text{text}} = -\frac{1}{B}\sum_{t=1}^{T} \log\, p(y_t \mid y_{1:t-1}, \mathbf{x})
\label{eq:ce_loss}
\end{equation}
where $T$ is the sequence length, $y_t$ is the target token at time step $t$, and $B$ is the batch size.

\subsection{Content Prompt Module}

To exploit the thematically structured nature of Dongba paintings, we introduce a content prompt module positioned after the encoder. This module maps the image feature vector $\mathbf{e}_x$ to a set of culture-aware content labels---such as \emph{deity}, \emph{Shu deity}, \emph{ghost}, \emph{archery}, \emph{fishing}, or \emph{flute playing}---via a fully connected layer followed by a softmax normalization:
\begin{equation}
S_i = \frac{\exp(e_i)}{\sum_{k=1}^{n} \exp(e_k)}
\label{eq:softmax_prompt}
\end{equation}
where $S_i$ is the normalized probability for the $i$-th category, $e_i$ and $e_k$ are the raw scores for the $i$-th and $k$-th categories, and $n$ is the total number of prompt categories. The predicted label corresponds to $\arg\max_i S_i$.

The predicted content label is then used to construct a prompt text $P$ in the form:
\begin{equation}
P = \text{``This is a Dongba painting about } [X_1, X_2, \ldots, X_m] \; \{b\}\text{''}
\label{eq:prompt_template}
\end{equation}
where $X_1, \ldots, X_m$ are learnable prompt vectors, and $b$ encodes the subject and action information derived from the image. The prompt text $P$ is tokenized into a prompt sequence $\mathbf{e}_p$ and concatenated with the caption text before being fed to the decoder.

To train the content prompt module, we minimize the cross-entropy between the predicted and ground-truth prompt labels:
\begin{equation}
\mathcal{L}_{\text{prompt}} = -\frac{1}{B}\sum_{i=1}^{B} \log\, p(y_i \mid x_i)
\label{eq:prompt_loss}
\end{equation}
where $y_i$ is the true content label and $x_i$ is the input feature for the $i$-th sample.

\subsection{Visual Semantic-Generation Fusion Loss}

To generate content-centric descriptions that are tightly aligned with the image, we design a composite loss that jointly optimizes the prompt prediction and the text generation objectives:
\begin{equation}
\mathcal{L}_{\text{fusion}} = \alpha \, \mathcal{L}_{\text{text}} + \lambda \, \mathcal{L}_{\text{prompt}}
\label{eq:fusion_loss}
\end{equation}
where $\alpha$ and $\lambda$ are weighting hyperparameters that control the relative contribution of the text generation loss and the prompt classification loss, respectively. By co-optimizing both objectives, the model is encouraged to extract culturally salient visual features through the encoder, while the decoder learns to produce descriptions that are both linguistically fluent and semantically faithful to the Dongba painting content.

\section{Experiments}

\subsection{Dataset}

Dongba paintings encompass a diverse array of artistic forms, including scripture scroll paintings, wooden tablet paintings, card paintings, and hanging scroll paintings. Following Qian et al.~\cite{qian2020simulation}, we collect authentic Naxi Dongba paintings and organize them into seven thematic categories: deity and spirit, hell ghost and demon, bird and beast, flora, horseback riding and fishing, music and dance, and religious pattern. Each image is annotated with a textual description that captures both the visual content and its cultural significance within the Dongba tradition. For example, animals in Dongba paintings are commonly depicted as divine mounts or mythologically significant creatures: the yak symbolizes valor and serves as the guardian beast of the Dongba altar; the white crane is an auspicious bird and a messenger of love; and the white bat, renowned for its cleverness, acts as a divine emissary in Naxi mythology, riding upon a sacred eagle to the heavenly realm to retrieve divinatory texts. Religious patterns frequently feature the Eight Treasures of the Naxi, including the purification vase (symbolizing fortune and spiritual purity), the endless knot (representing unity and harmony), and the twin fish (denoting vitality and freedom).

\subsection{Data Augmentation}

Deep learning models typically require large-scale training data, yet authentic Dongba paintings are scarce and their descriptive annotations are limited. To mitigate overfitting, we apply three standard categories of data augmentation~\cite{shorten2019survey}: (1)~\emph{geometric transformations} (rotation, flipping, cropping), which help the model learn features invariant to viewing angle; (2)~\emph{noise injection} (random Gaussian noise), which forces the model to focus on salient descriptive features rather than incidental noise; and (3)~\emph{pixel operations} (color jittering, sharpening), which simulate the natural degradation of Dongba paintings due to aging and oxidation. After augmentation, the dataset is expanded to 9{}408 images.

\subsection{Experimental Setup}

All experiments are conducted using the TensorFlow and Keras deep learning frameworks on an NVIDIA RTX 2080Ti GPU. The encoder is MobileNetV2 pre-trained on ImageNet~\cite{deng2009imagenet}~\cite{deng2009imagenet}, and the decoder consists of 10 Transformer coding layers initialized with pre-trained BERT weights. The content prompt module is trained jointly with the decoder. The batch size is set to 8, the initial learning rate is $1 \times 10^{-5}$, and the optimizer is AdamW~\cite{loshchilov2017decoupled} with a weight decay rate of 0.8 every 8 epochs.

\subsection{Evaluation Metrics}

We adopt four widely used image captioning evaluation metrics:

\paragraph{BLEU}~\cite{papineni2002bleu} evaluates the precision of $n$-gram matches between the generated and reference captions:
\begin{equation}
f_{\text{BLEU}} = \text{BP} \times \exp\!\left(\sum_{n=1}^{N} w_n \log P_n\right)
\end{equation}
where $P_n$ is the modified $n$-gram precision and BP is the brevity penalty.

\paragraph{METEOR}~\cite{banerjee2005meteor} assesses lexical matching, morphological variation, and semantic similarity:
\begin{equation}
f_{\text{METEOR}} = (1 - \text{Pen}) \times \frac{P_m R_m}{\eta P_m + (1-\eta) R_m}
\end{equation}

\paragraph{ROUGE}~\cite{lin2004rouge} measures the recall of the longest common subsequence between the generated and reference texts:
\begin{equation}
f_{\text{ROUGE\_L}} = \frac{(1 + \mu^2) R_{\text{lcs}} P_{\text{lcs}}}{R_{\text{lcs}} + \mu^2 P_{\text{lcs}}}
\end{equation}

\paragraph{CIDEr}~\cite{vedantam2015cider} computes TF-IDF weighted $n$-gram similarity between the generated caption and multiple reference captions:
\begin{equation}
f_{\text{CIDEr}}(c_i, S_i) = \frac{1}{J}\sum_{j=1}^{J}\sum_{n=1}^{N} \frac{\mathbf{g}_n(c_i) \cdot \mathbf{g}_n(s_{ij})}{\|\mathbf{g}_n(c_i)\| \, \|\mathbf{g}_n(s_{ij})\|}
\end{equation}

\subsection{Effect of Different Encoders}

To investigate the impact of the visual encoder on captioning quality, we compare five encoder architectures while keeping all other model components fixed. Table~\ref{tab:encoder} presents the results.

\begin{table}[t]
\centering
\caption{Comparison of image captioning metrics with different encoders. Bold values indicate the best result in each column.}
\label{tab:encoder}
\small
\begin{tabular}{lccccccc}
\toprule
\textbf{Encoder} & \textbf{Params} & \textbf{B-1} & \textbf{B-2} & \textbf{B-3} & \textbf{B-4} & \textbf{MET.} & \textbf{ROU.} \\
& (M) & & & & & & \\
\midrule
MobileNetV2 & 8.73 & \textbf{0.603} & \textbf{0.426} & \textbf{0.317} & 0.246 & \textbf{0.256} & \textbf{0.403} \\
EfficientNetB0 & 15.55 & 0.588 & 0.410 & 0.311 & \textbf{0.247} & 0.251 & 0.399 \\
DenseNet121 & 26.93 & 0.581 & 0.408 & 0.313 & 0.249 & 0.252 & 0.396 \\
VGG16 & 56.17 & 0.554 & 0.352 & 0.251 & 0.187 & 0.223 & 0.365 \\
ResNet-50 & 90.16 & 0.606 & 0.422 & 0.314 & 0.243 & 0.252 & 0.399 \\
\bottomrule
\end{tabular}
\vspace{0.2cm}

\begin{tabular}{lc}
\toprule
\textbf{Encoder} & \textbf{CIDEr} \\
\midrule
MobileNetV2 & \textbf{0.599} \\
EfficientNetB0 & 0.571 \\
DenseNet121 & 0.549 \\
VGG16 & 0.386 \\
ResNet-50 & 0.545 \\
\bottomrule
\end{tabular}
\end{table}

When MobileNetV2 is used as the encoder, the model achieves the best overall performance with only 8.73M parameters. Replacing the encoder with EfficientNetB0 or VGG16 increases the parameter count by 6.82M and 47.44M, respectively, yet all metrics decline to varying degrees. DenseNet121 yields a marginal improvement in BLEU-4 (+0.006) but reduces other metrics. ResNet-50~\cite{he2016deep}, despite having 90.16M parameters, achieves a slightly higher BLEU-1 (0.606) but produces lower scores on the remaining six metrics. These results confirm that MobileNetV2 provides an optimal balance between parameter efficiency and captioning quality for the Dongba painting domain.

\subsection{Effect of Decoder Depth}

To examine the influence of decoder depth on caption quality, we vary the number of coding layers (6, 8, 10, 12) while keeping the encoder and other components unchanged. As the number of layers increases from 6 to 10, all metrics improve steadily, indicating that additional layers enhance the model's ability to capture fine-grained visual details and generate diverse, coherent descriptions. However, beyond 10 layers, performance gains diminish and certain metrics begin to decline due to overfitting. We therefore select 10 coding layers as the default configuration.

\subsection{Comparison with State-of-the-Art Methods}

\subsubsection{Objective Evaluation}

Table~\ref{tab:comparison} compares PVGF-DPC against nine baseline models on the Dongba painting test set.

\begin{table*}[t]
\centering
\caption{Comparison of image captioning metrics between PVGF-DPC and baseline methods on the Dongba painting test set. Bold values indicate the best result in each column; ``--'' denotes missing or unreliable values.}
\label{tab:comparison}
\small
\begin{tabular}{lccccccc}
\toprule
\textbf{Model} & \textbf{BLEU-1} & \textbf{BLEU-2} & \textbf{BLEU-3} & \textbf{BLEU-4} & \textbf{METEOR} & \textbf{ROUGE} & \textbf{CIDEr} \\
\midrule
ATT + LSTM & 0.381 & 0.104 & 0.063 & 0.045 & 0.178 & 0.400 & 0.046 \\
ClipCap~\cite{mokady2021clipcap} & 0.497 & 0.209 & 0.139 & 0.104 & 0.239 & 0.142 & 0.472 \\
BLIP~\cite{li2022blip} & 0.365 & 0.102 & 0.054 & 0.030 & 0.180 & 0.361 & 0.026 \\
OFA~\cite{wang2022ofa} & 0.066 & -- & -- & -- & 0.251 & 0.191 & -- \\
DeCap~\cite{li2023decap} & 0.208 & -- & -- & -- & 0.104 & 0.280 & -- \\
CgT-GAN~\cite{yu2023cgtgan} & 0.372 & 0.288 & 0.091 & 0.033 & 0.148 & 0.299 & 0.038 \\
ViECap~\cite{fei2023transferable} & 0.457 & 0.285 & 0.185 & 0.127 & 0.148 & 0.308 & 0.183 \\
MeaCap~\cite{zeng2024meacap} & 0.119 & 0.056 & 0.026 & 0.011 & 0.072 & 0.185 & 0.014 \\
MacCap~\cite{qiu2024mining} & 0.186 & 0.123 & 0.079 & 0.056 & 0.083 & 0.245 & 0.012 \\
\midrule
\textbf{PVGF-DPC (Ours)} & \textbf{0.603} & \textbf{0.426} & \textbf{0.317} & \textbf{0.246} & \textbf{0.256} & \textbf{0.403} & \textbf{0.599} \\
\bottomrule
\end{tabular}
\end{table*}

PVGF-DPC achieves the highest scores across all seven metrics. In terms of accuracy and fluency, BLEU-1 through BLEU-4 surpass the second-ranked ClipCap by margins of 0.106, 0.217, 0.178, and 0.142, respectively. The CIDEr score of 0.599 exceeds that of the second-ranked ViECap by 0.416, demonstrating markedly superior relevance and diversity. The METEOR score of 0.256 surpasses the second-ranked OFA by 0.005, indicating better word-level and semantic alignment. These substantial improvements highlight the effectiveness of the content prompt module and the visual semantic-generation fusion loss in directing the model toward culturally accurate and linguistically fluent Dongba painting descriptions.

\subsubsection{Subjective Evaluation}

To provide a more intuitive assessment, we present qualitative comparisons of generated captions for representative Dongba paintings. In the first example, depicting a white bat, PVGF-DPC not only correctly identifies the subject and its action (``the white bat flapping its wings'') but also describes its mythological role in Dongba culture (``in Dongba mythology, Hengying-Jinbaipan [the white bat] serves as a divine messenger, riding upon the sacred eagle Aluoxiugong to the heavenly realm to retrieve divinatory texts''). In contrast, BLIP, ViECap, and MacCap misidentify the subject as a white crane, a white dragon, or a generic bird, respectively. In the second example, featuring a Shu deity, PVGF-DPC accurately describes the costume (``wearing yellow garments''), the headpiece (``adorned with a jeweled crown''), and the traditional iconography (``human body with a serpentine tail, solemn and majestic''). In the third example, depicting a purification vase, PVGF-DPC provides detailed visual descriptions (``a blue purification vase with green ribbon embellishments'') along with its cultural significance within the Dongba tradition (``symbolizing fortune, prosperity, and spiritual purity'').

These qualitative results confirm that the content prompt module successfully steers the decoder toward culturally faithful descriptions, while the fusion loss ensures that the generated text is visually grounded and semantically accurate.

\subsection{Ablation Study}

To quantify the individual contributions of the content prompt module and the visual semantic-generation fusion loss, we conduct ablation experiments with three configurations: (1)~DBC (Dongba captioning): the base encoder--decoder model without the fusion loss; (2)~VGF-DPC: employing only the fusion loss without the content prompt module; and (3)~PVGF-DPC: the full model with both components. Results are presented in Table~\ref{tab:ablation}.

\begin{table}[t]
\centering
\caption{Ablation study results. Bold values indicate the best result in each column.}
\label{tab:ablation}
\small
\begin{tabular}{lccccccc}
\toprule
\textbf{Model} & \textbf{B-1} & \textbf{B-2} & \textbf{B-3} & \textbf{B-4} & \textbf{MET.} & \textbf{ROU.} & \textbf{CIDEr} \\
\midrule
DBC & 0.575 & 0.392 & 0.291 & 0.227 & 0.247 & 0.392 & 0.526 \\
VGF-DPC & 0.591 & 0.407 & 0.305 & 0.239 & 0.247 & 0.401 & 0.489 \\
\textbf{PVGF-DPC} & \textbf{0.603} & \textbf{0.426} & \textbf{0.317} & \textbf{0.246} & \textbf{0.256} & \textbf{0.403} & \textbf{0.599} \\
\bottomrule
\end{tabular}
\end{table}

Compared to the DBC baseline, the full PVGF-DPC model improves BLEU-1 by 0.028, ROUGE by 0.011, and CIDEr by 0.073. The VGF-DPC variant, which includes the fusion loss but omits the prompt module, shows intermediate performance, confirming that both components contribute complementary benefits: the fusion loss strengthens the encoder's ability to extract thematically relevant visual features, while the content prompt module provides explicit cultural guidance that enhances the decoder's descriptive accuracy. The CIDEr improvement of 0.110 from VGF-DPC to PVGF-DPC is particularly notable, indicating that the prompt module substantially enriches the diversity and cultural specificity of the generated descriptions.

\section{Conclusion}

This paper presents PVGF-DPC, a prompt-guided captioning framework for Dongba paintings that addresses the twin challenges of data scarcity and domain shift through the coordinated action of a content prompt module and a visual semantic-generation fusion loss. The content prompt module maps image features to culture-aware thematic labels and constructs post-prompts that supply the Transformer decoder with explicit cultural context, thereby mitigating hallucination and enhancing the relevance of generated descriptions to the distinctive iconography of Naxi Dongba art. The visual semantic-generation fusion loss jointly optimizes the prompt classification and caption generation objectives, encouraging the MobileNetV2 encoder to extract culturally salient features and the BERT-initialized decoder to produce fluent, content-centric descriptions. Experimental evaluation on a dedicated Dongba painting dataset of 9{}408 augmented images demonstrates that PVGF-DPC attains BLEU-1/2/3/4 scores of 0.603/0.426/0.317/0.246, a METEOR score of 0.256, a ROUGE score of 0.403, and a CIDEr score of 0.599, substantially outperforming competitive baselines---including BLIP, ViECap, MacCap, ClipCap, and OFA---in both objective metrics and subjective cultural expressiveness; ablation experiments further confirm that the content prompt module and the fusion loss provide complementary benefits, with their combination yielding a CIDEr improvement of 0.073 over the base encoder--decoder model. Future work will explore more sophisticated prompt designs that capture finer-grained cultural attributes, expand the Dongba painting dataset in both scale and thematic diversity, and investigate the transferability of the proposed framework to other cultural heritage visual domains.

\bibliographystyle{plainnat}
\bibliography{references}

\end{document}